\documentclass[conference,onecolumn,a4paper,10pt]{IEEEtranAdapt}
\usepackage[margin=25mm,tmargin=25mm,bmargin=25mm,headsep=25pt,voffset=6pt,footskip=30pt]{geometry}
\usepackage{cite}
\usepackage{amsmath,amssymb,amsfonts}
\usepackage{algorithmic}
\usepackage{graphicx}
\usepackage{textcomp}
\usepackage{xcolor}
\usepackage{tabularx}
\usepackage{caption}
\usepackage{url}
\usepackage{listings}
\usepackage{wrapfig,blindtext}
\usepackage{subfigure}
\usepackage[export]{adjustbox}

\newif\ifarxiv
\arxivtrue

\usepackage{fancyhdr}

\ifarxiv
    \usepackage[numbers]{natbib}
\else
    \usepackage{natbib}
\fi

\usepackage[
colorlinks=true,
linkcolor=blue,
urlcolor=blue,
citecolor=blue,
anchorcolor=blue
]{hyperref}

\ifarxiv
    \newcommand\boldarial{\sffamily\bfseries}
    \newcommand\narrowarial{\sffamily\bfseries}
    \newcommand\narrowitalicarial{\sffamily\bfseries}
\else
    \newfontfamily\boldarial[Path=./font/]{Arial Rounded MT Bold.ttf}
    \newfontfamily\narrowarial[Path=./font/]{Arial Narrow.ttf}
    \newfontfamily\narrowitalicarial[Path=./font/]{Arial Narrow Italic.ttf}
\fi

\ifarxiv
\else
    \definecolor{bleudefrance}{rgb}{0.19, 0.55, 0.91}
    
\fi

\renewcommand{\thesection}{\arabic{section}}
\renewcommand{\thesubsectiondis}{\mbox{\thesection.\arabic{subsection}}}
\renewcommand{\thesubsubsectiondis}{\mbox{\thesubsectiondis.\arabic{subsubsection}}}

\renewcommand{\thetable}{\arabic{table}}
\newcommand{\opensi}{OpenSI-CoSMIC}

\makeatletter
\renewcommand{\maketitle}{\bgroup\setlength{\parindent}{0pt}
\begin{flushleft}
  {\fontsize{16pt}{20pt}\selectfont{\boldarial{\@title}}}\\
  \vspace{12pt}
  \@author
  \vspace{6pt}
\end{flushleft}\egroup
}

\renewenvironment{abstract}
{\par\vspace{6pt}{\fontsize{14pt}{18pt}\selectfont\boldarial{\abstractname}}\vspace{6pt}\\}
{\par\medskip}

\renewenvironment{IEEEkeywords}
{\par\textbf{\IEEEkeywordsname}}
{\par\medskip}

\renewenvironment{section}{\@startsection{section}{1}{\z@}{1.5ex plus 1.5ex minus 0.5ex}%
{0.7ex plus 1ex minus 0ex}{\Large\raggedright\boldarial}}%

\renewenvironment{subsection}{\@startsection{subsection}{2}{\z@}{1.5ex plus 1.5ex minus 0.5ex}%
{0.7ex plus .5ex minus 0ex}{\large\boldarial}}%
\makeatother

\begin{document}

% =================================================================================================================================
% Do not change these except the author information
\pagestyle{fancy}

% \ifanonymous
% \title{Unleashing Artificial Cognition: Integrating Multiple AI Systems\\
% \vspace{7pt}
% \fontsize{12pt}{15pt}\selectfont{Full research paper}}

% \author{
% {\fontsize{12pt}{15pt}\selectfont\boldarial{<removed for refereeing>}} \\
% {\fontsize{10pt}{13pt}\selectfont{<removed for refereeing>}} \\
% {\fontsize{10pt}{13pt}\selectfont{Email: <removed for refereeing>@<removed for refereeing>.edu.au}}
% \vspace{8pt}

% {\fontsize{12pt}{15pt}\selectfont\boldarial{<removed for refereeing>}} \\
% {\fontsize{10pt}{13pt}\selectfont{<removed for refereeing>}} \\
% {\fontsize{10pt}{13pt}\selectfont{Email: <removed for refereeing>@<removed for refereeing>.edu.au}}
% \vspace{8pt}

% {\fontsize{12pt}{15pt}\selectfont\boldarial{<removed for refereeing>}} \\
% {\fontsize{10pt}{13pt}\selectfont{<removed for refereeing>}} \\
% {\fontsize{10pt}{13pt}\selectfont{Email: <removed for refereeing>@<removed for refereeing>.edu.au}}
% \vspace{8pt}

% {\fontsize{12pt}{15pt}\selectfont\boldarial{<removed for refereeing>}} \\
% {\fontsize{10pt}{13pt}\selectfont{<removed for refereeing>}} \\
% {\fontsize{10pt}{13pt}\selectfont{Email: <removed for refereeing>@<removed for refereeing>.edu.au}}
% \vspace{8pt}

% {\fontsize{12pt}{15pt}\selectfont\boldarial{<removed for refereeing>}} \\
% {\fontsize{10pt}{13pt}\selectfont{<removed for refereeing>}} \\
% {\fontsize{10pt}{13pt}\selectfont{Email: <removed for refereeing>@<removed for refereeing>.edu.au}}
% \vspace{6pt}
% }
% \else
\title{Unleashing Artificial Cognition: Integrating Multiple AI Systems
\vspace{8pt}}

\author{
{\fontsize{12pt}{15pt}\selectfont\boldarial{Muntasir Adnan}} \\
{\fontsize{10pt}{13pt}\selectfont{Open Source Institute, Faculty of Science and Technology, University of Canberra}} \\
{\fontsize{10pt}{13pt}\selectfont{Email: adnan.adnan@canberra.edu.au}}
\vspace{8pt}

{\fontsize{12pt}{15pt}\selectfont\boldarial{Buddhi Gamage}} \\
{\fontsize{10pt}{13pt}\selectfont{Collaborative Robotics Lab, Faculty of Science and Technology, University of Canberra}} \\
{\fontsize{10pt}{13pt}\selectfont{Email: buddhi.gamage@canberra.edu.au}}
\vspace{8pt}

{\fontsize{12pt}{15pt}\selectfont\boldarial{Zhiwei Xu}} \\
{\fontsize{10pt}{13pt}\selectfont{Open Source Institute, Faculty of Science and Technology, University of Canberra}} \\
{\fontsize{10pt}{13pt}\selectfont{Email: danny.xu@canberra.edu.au}}
\vspace{8pt}

{\fontsize{12pt}{15pt}\selectfont\boldarial{Damith Herath}} \\
{\fontsize{10pt}{13pt}\selectfont{Collaborative Robotics Lab, Faculty of Science and Technology, University of Canberra}} \\
{\fontsize{10pt}{13pt}\selectfont{Email: damith.herath@canberra.edu.au}}
\vspace{8pt}

{\fontsize{12pt}{15pt}\selectfont\boldarial{Carlos C. N. Kuhn}} \\
{\fontsize{10pt}{13pt}\selectfont{Open Source Institute, Faculty of Science and Technology, University of Canberra}} \\
{\fontsize{10pt}{13pt}\selectfont{Email: carlos.noschangkuhn@canberra.edu.au}}
\vspace{6pt}
}
% \fi

% =================================================================================================================================
\maketitle

\begin{abstract}

In this study, we present an innovative fusion of language models and query analysis techniques to unlock cognition in artificial intelligence. The introduced open-source AI system seamlessly integrates a Chess engine with a language model, enabling it to predict moves and provide strategic explanations. Leveraging a vector database to achieve retrievable answer generation, our AI system elucidates its decision-making process, bridging the gap between raw computation and human-like understanding.
Our choice of Chess as the demonstration environment underscores the versatility of our approach. Beyond Chess, our system holds promise for diverse applications, from medical diagnostics to financial forecasting.
Our AI system is available at \href{https://github.com/TheOpenSI/CoSMIC.git}{https://github.com/TheOpenSI/CoSMIC.git}.
\end{abstract}


\newpage

% =================================================================================================================================
\section{Introduction}
\label{sec:introduction}

Artificial Intelligence (AI) systems have achieved remarkable feats in specialized areas such as image recognition and natural language processing \citep{patil2024review,khurana2023natural, zhao2023brain}. Despite these advancements, individual AI models typically excel in isolated tasks and lack general cognition abilities, leading to Artificial General Intelligence (AGI)~\citep{sonko2024critical}. This fragmentation restricts their potential for broader and more generalized applications requiring seamless interaction of multiple cognitive functions.

Human cognition is marked by adaptability, creativity, and emotional intelligence, guided by goals, norms, and social/ethical considerations~\citep{siemens2022human}. In contrast, artificial cognition involves simulating these processes in machines to perform tasks autonomously~\citep{qu2024integration}. Studies have highlighted the strengths and limitations of human and artificial cognition, emphasizing the need for understanding their difference for effective human-AI collaboration~\citep{korteling2021human}.

The Turing Test, introduced by Alan Turing~\citep{turing1950mind}, posits that a machine can be considered intelligent if it can carry on a conversation indistinguishable from a human. Despite its historical significance, the Turing Test has notable limitations. It is anthropocentric, assuming human-like conversation as the definitive marker of intelligence, thereby excluding other forms of intelligence like complex problem-solving or creative pattern recognition. Critics, including Turing, have argued that pre-programmed responses could deceive the interrogator, undermining the test's ability to assess cognitive abilities~\citep{halpern2006trouble, searle1980minds}. Additionally, the Turing Test lacks granularity in evaluating cognition, as it does not assess various cognitive abilities such as attention, memory, learning,  and reasoning, nor does it compare AI's cognitive stages to human levels~\citep{piaget1971theory}.

Evaluating cognition in AI involves assessing the system's ability to perform tasks requiring intelligence and adaptation to various situations. This includes simulating human-like cognitive processes to enable socially intelligent and adaptive interactions with humans~\citep{vernon2014artificial, vernon2016desiderata, kotseruba202040}. By incorporating specific tasks that assess the mentioned cognitive qualities, we aim to create a more comprehensive assessment strategy for AI cognition, offering insights into the strengths and weaknesses of AI systems.

In the book ``Cognitive Robotics", \citet{cangelosi2022cognitive} discuss eight cognitive abilities, drawing on the work of~\citep{kotseruba202040}, who examine seven essential cognitive abilities: perception, attention mechanisms, action selection, memory, learning, reasoning, and meta-reasoning.~\citet{vernon2007survey} add anticipation to this list.

Following these ideas, in this study centred around chess, we identified five cognitive qualities relevant to chess players for making decisions during game-play.
%six cognitive abilities to demonstrate cognition. 
% While learning is another crucial ability in cognition and essential for enhancing current cognitive levels, it is not the focus of our paper. Similarly, meta-reasoning, which involves the ability to reason about one’s own reasoning processes, is not considered in our evaluation. Action selection is considered as a part of anticipation and will be further discussed in the methodology under the scoring mechanism. Instead, we aim to provide a score based on these five qualities to evaluate cognitive performance in this context.
The cognitive qualities we focus on are:

\begin{itemize}
    \item \textbf{Perception}. The ability to interpret and understand sensory information from the environment.
    \item \textbf{Memory.} The capability to store, retain, and retrieve information.
    \item \textbf{Attention.} The skill of focusing on relevant stimuli while filtering out distractions.
    \item \textbf{Reasoning.} The ability to draw logical inferences and conclusions from available information.
    \item \textbf{Anticipation.} The capability to predict future events or outcomes based on current information and past experiences.
\end{itemize}
 
In this work, we introduce an open-source AI system, namely Open Source Institute-Cognitive System of Machine Intelligent Computing (\opensi).
It focuses on developing the initial requirements for an AI system to achieve higher cognition levels in a closed environment. We present a systematic way to evaluate the cognitive capabilities of our integrated system. We show that individual models may exhibit cognitive qualities independently, and their integration can lead to the emergence of cognitive behaviours comparable to humans.
% ---------------------------------------------------------------------------------------------------------------------------------
\section{Methodology}

% ---------------------------------------------------------------------------------------------------------------------------------
\subsection{Proposed System for Demonstrating Cognitive Abilities}

% Our proposed system integrates multiple AI models and tools, each specializing in a different cognitive function, to create a cohesive framework that demonstrates comprehensive cognitive abilities. This system includes models for the aforementioned cognitive qualities.
Our proposed system integrates multiple AI models and tools, each specialising in different aforementioned cognitive qualities. By integrating these tools, we aim to enable the system to perform complex tasks that require the interplay of multiple cognitive functions, thus exhibiting cognition.

While constituting a mainstream and \textit{demonstrably effective} set, the employed techniques are acknowledged to be limited. More advanced fine-tuning, Retrieval-Augmented Generation (RAG)~\citep{lewis2020retrieval}, and Retrieval-Augmented Fine-Tuning (RAFT)~\citep{zhang2024raft}, may offer \citep{gekhman2024does} further performance enhancements. Nonetheless, this study combines these mainstream technologies to assess the feasibility and potential for introducing human-like cognitive capabilities within an AI system. The proposed system encompasses a range of services that the agents can decide to employ. To evaluate its efficacy, the system is designed to experiment with several Large Language Models (LLMs). The system is comprised of the following components:  
\begin{itemize}
    \item A query analyser service.
    \item Base LLM or a fine-tuned LLM using Parameter-Efficient Fine-Tuning's (PEFT)~\citep{xu2023parameter} and Low-Rank Adaptation (LoRA)~\citep{hu2021lora}.
    \item An external knowledge source facilitated by a Faiss vector database and RAG capability.
    \item A chess engine service powered by Stockfish.
    \item A vector database update service that allows real-time information updates.
\end{itemize}

\begin{figure}[!ht]
    \centering
    \includegraphics[width=\linewidth]{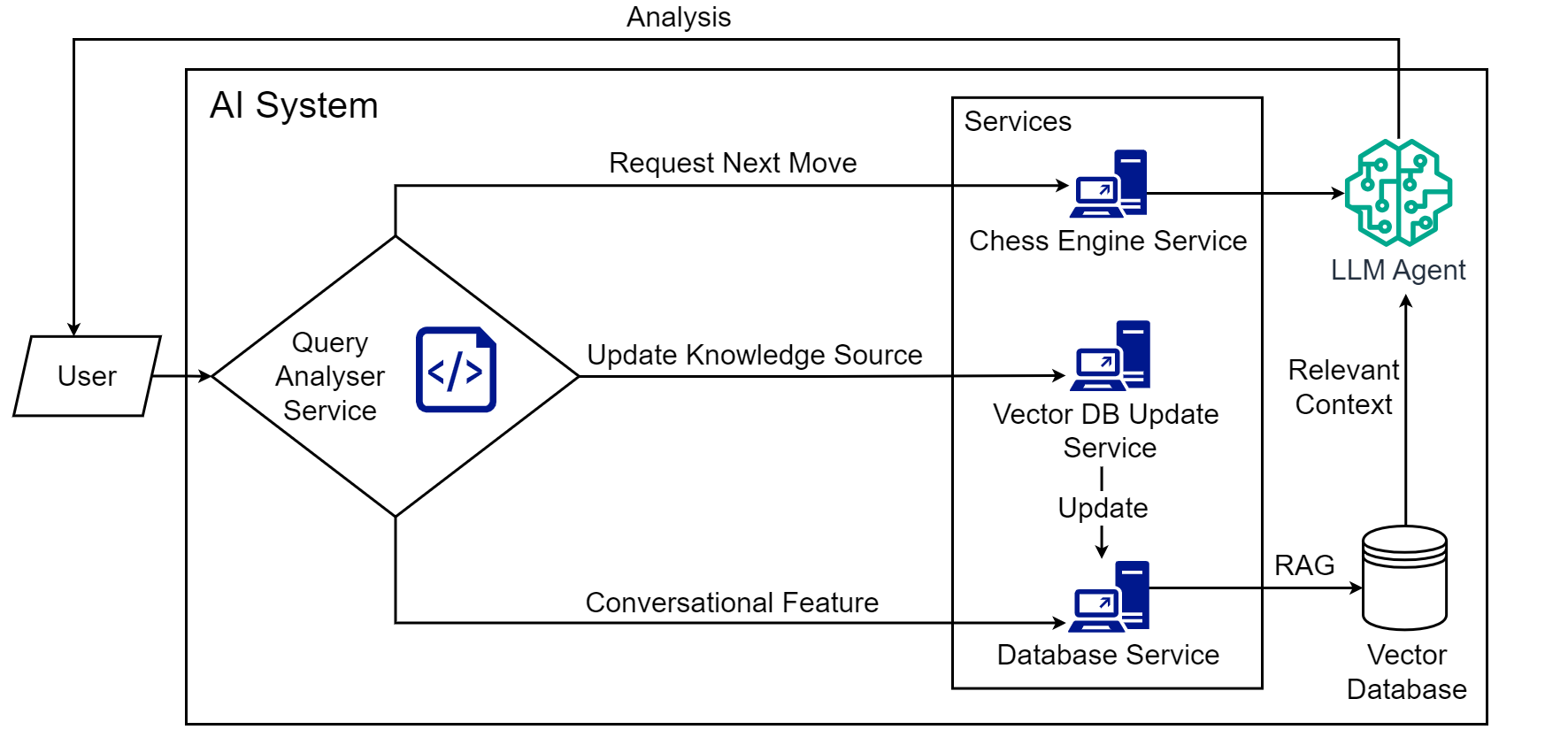}
    \caption{System Architecture and Data Flow}
    \label{System_Architecture}
\end{figure}

The components mentioned above work together to achieve cognitive qualities within the system. Figure \ref{System_Architecture} illustrates this collaboration in detail, depicting the system architecture.

\textbf{Fine-tuning.} For fine-tuning, we leverage an instruction tuning~\citep{wang2022instruction, wei2021finetuned_instruction} methodology.
The base model is Mistral 7B, chosen for its balance of performance, efficiency and size.

To promote slow and deliberative reasoning in a small student model, we employ a teacher-student learning paradigm~\citep{mitra2023orca}. OpenAI's GPT-4o served as the teacher model, and we interacted with it using specific system prompts designed to elicit a deliberate step-by-step reasoning approach while generating responses. These outputs form a core component of the training data. Conversely, during fine-tuning the student model, a generic system prompt is employed. This generic prompt, used for future inference across all models, would allow the student model to leverage its learned reasoning skills as its default behaviour. This is because the pre-trained teacher model, due to its size and capacity, may provide answers directly, while the small student model will require a more deliberative approach to provide similar answers~\citep{mitra2023orca}. This distinction in prompt engineering transfers the generalised reasoning abilities from the teacher model to the student model.

Additionally, a dataset of chess games with detailed and step-by-step reasoning annotations for each move, akin to a ``chain-of-thought" approach~\citep{wei2022chain}, is constructed~\citep{zhang2024raft}.
% \ifanonymous
% and available at <removed for refereeing>.
% \else
% .
% \fi
This custom chess reasoning dataset is further supplemented by the publicly available Kaggle Lichess dataset to provide broader coverage of chess scenarios.

The finetuning training dataset included the following tasks:
\begin{itemize}
    \item \textbf{Chess Move Analysis.} Explain the rationale for a chess move prediction with reference analysis from OpenAI.
    \item \textbf{Next Move Prediction.} Predict the best next move played by a human opponent with samples from Kaggle Lichess.
    \item \textbf{Game Winner Prediction.} Predict the winner of a chess game based on the current state with samples from Kaggle Lichess and OpenAI.
    \item \textbf{Piece Capture Analysis.} Analyse potential captures based on a list of moves in Algebraic Notation using custom scripts.
    \item \textbf{FEN Parsing and Reasoning.} Reason from a position represented in Forsyth-Edwards Notation (FEN) format~\citep{PGNSpec} using custom scripts.
    \item \textbf{Retrieval-Augmented Fine-Tuning dataset.} Analyse the provided contexts and use the most relevant context to generate an answer, following work of \citep{zhang2024raft}
\end{itemize}

% This diverse training regime aimed to develop the model's ability to ``think" in different chess-related scenarios and foster transferable reasoning capabilities.
The diversity of training data aims to enhance the model's ability to ``think" in various chess scenarios and develop transferable reasoning capabilities. Additionally, we introduce new knowledge to the model through fine-tuning, enabling it to understand and utilise FEN. Although the model successfully demonstrates this new capability, the new knowledge increases its propensity for hallucinations when confronted with zero-shot tasks, which agrees with the observations from \citep{gekhman2024does}.

To optimize computational efficiency during fine-tuning, we employ several techniques. The base Mistral 7B model is quantized to a 4-bit representation~\citep{dettmers2024qlora}, significantly reducing the computational complexity regarding memory and running time. Furthermore, instead of fully fine-tuning the base model, we leverage Low-Rank Adaptation (LoRA) to fine-tune a smaller adapter module. This approach further reduces the computational overhead for fine-tuning. The quantization configuration is consistent across all models utilising Hugging Face Transformers~\citep{wolf2019huggingface}.

For inference, we use the finetuned PEFT model, which involves adding the fine-tuned LoRA adapters to the quantized base model. Interestingly, we observe behavioural discrepancies between the fine-tuned model when adding the LoRA adapter (as a PEFT model) and fully merging the adapter into the base Mistral 7B model. While the PEFT model performs as expected during inference, a more detailed investigation into this discrepancy is warranted for future studies.

Additionally, to assess the effectiveness of the proposed fine-tuning methodology for comparisons, our system is evaluated with base models from other LLM families, including GPT-4o, GPT-3.5 Turbo, Gemma 7B Instruct, and Mistral 7B Instruct. The results will be presented in the experiments section.

\textbf{Retrieval-Augmented Generation (RAG).} We adopt RAG~\citep{lewis2020retrieval} in our framework to leverage external knowledge sources stored in a Faiss vector data store~\citep{douze2024faiss}. An embedding model plays a crucial role in transforming both textual information from the knowledge source and user queries into high-dimensional vectors. This enables efficient similarity search using distance measurement~\citep{dokmanic2015euclidean} between the query vector and vectors representing data points in the Faiss database, where the smallest distance indicates the most similar context to the query vector. The \texttt{similarity\_search\_with\_score} from LangChain \citep{chase2023langchain} facilitates the retrieval of the most relevant contexts based on these distances.

A pre-defined similarity threshold filters retrieve information to extract highly relevant knowledge for reasoning. This is crucial as we aimed to assess the system's ability to exhibit reasoning and attention across diverse scenarios, potentially extending beyond the immediate domain of chess.

All base LLMs used in the system leverage the same embedding model \texttt{embedding\_model\_name} retrieved from the Hugging Face library for seamless integration with the Faiss vector store, which works seamlessly for both pipelines using LangChain and the Llama indexing~\citep{Liu_LlamaIndex_2022} library. However, combining the fine-tuned model requires slightly more advanced techniques since Llama Indexing currently lacks support for PEFT models. Additionally, the fine-tuned model necessitates further training to leverage retrieved contexts effectively. To address these concerns, we use RAFT and provide our strategy below.

\textbf{Retrieval-Augmented Fine-Tuning (RAFT).} The base models perform optimally within the RAG framework, effectively utilising the provided contexts. However, the fine-tuned model exhibits confusion when presented with the extracted contexts from the documents. Although the model demonstrates its capability to tackle various problem-solving tasks, including novel queries for piece-capture and game-analysis problems in a zero-shot manner, it struggles to deal with entirely novel contexts, leading to hallucinations in response. Consequently, further fine-tuning is necessary to instruct the model on effectively leveraging the provided contexts~\citep{zhang2024raft}.

We adopt the methodology outlined in~\citep{zhang2024raft}, where the authors train an LLM to concentrate on the most relevant context by inducing a slow reasoning analysis of all given contexts, similar to the Chain of Thought (CoT) approach. The resultant dataset comprises a user query, relevant data points, and a CoT-styled analysis of the contexts as the answer, thus instructing the model on how to analyse and utilise the provided contexts. We also include training examples where the model is expected to disregard all provided contexts upon analysis and instead rely on its pre-trained knowledge to respond to the query. An example of this dataset is provided below. 

\begin{lstlisting}[language={}, frame=single, basicstyle=\small]
Who are Tardo and Peo in {DISQUALIFIED} by Charles L. Fontenay?
-----
To determine who Tardo and Peo are in {DISQUALIFIED} by Charles L. Fontenay, we can
look at the provided context. 

1. The context mentions that Tardo, the Solar Council's Planetary Aid agent, and his
companion, Peo, were taken to the castle. This indicates that Tardo is the Solar
Council's Planetary Aid agent, and Peo is his companion.

Therefore, based on the context provided, Tardo is the Solar Council's Planetary Aid
agent, and Peo is his companion.

<ANSWER>: Tardo is the Solar Council's Planetary Aid agent, and Peo is his companion.
\end{lstlisting}

While this approach successfully introduces the fine-tuned model to consider the retrieved contexts, we believe this specific methodology warrants a dedicated study to enhance the RAG capability among various problem-solving tasks for large language models (LLMs).

\textbf{Chess Engine Service.} The chess engine service is an integral component of our system, leveraging the capabilities of the Stockfish engine~\citep{romstad2008stockfish} to evaluate and demonstrate its cognitive abilities within the domain of chess. Particularly, Stockfish is a highly advanced and open-source chess engine renowned for its robust performance and ability to analyse positions and predict potential moves. In our system, the chess engine service interacts seamlessly with the LLM, providing real-time move suggestions. This integration enables the AI system to delegate tasks to a more capable tool and articulate the rationale of each move, thereby showcasing its reasoning and anticipation capabilities.

\textbf{Vector Database Update Service.} The vector database update service plays a pivotal role in maintaining and enhancing the system's memory and knowledge retrieval capabilities. This service is designed to update the Faiss~\citep{douze2024faiss} vector database in real time, ensuring that the AI system can access the most current and relevant information. This dynamic updating mechanism is critical for the RAG framework, which relies on accurate and up-to-date information to provide contextually relevant responses. The vector database update service ensures that the AI model can learn from its experiences, store important information, and retrieve it when needed, thereby enhancing its cognitive abilities related to memory. This service underscores the system's capacity for continuous improvement and adaptation, essential traits for demonstrating artificial cognition in varied and evolving scenarios.

% =================================================================================================================================
\subsection{Scoring Mechanism}
This section explores how to assess an AI System's cognitive qualities in a closed environment; in this case, we are using Chess as the closed environment.

To quantify the cognition capability of our AI system, we design a scoring mechanism with the aforementioned 5 qualities from the perspectives of the agent's environment understanding, information processing, and solution provision.
% In below, we will define the i$^{\text{th}}$ quality for all $i \in \mathcal{Q}$, where $\mathcal{Q}$ is the set of qualities.
%
To this end, we have developed a Question-and-Answer (Q\&A) testing system, wherein the list of questions was meticulously curated to consider each quality. It is important to note that the dataset used in this study for the particular chess domain is not exhaustive; the questions can be generalized to other domains and the AI system's cognition level~\citep{piaget1971theory}.

The evaluation of each quality is based on statistics of sufficient test samples.
We provide the details and their corresponding assessment criteria below. These criteria form the basis of our evaluation, ensuring a thorough and systematic assessment of AI’s cognitive capabilities.
Standard Algebraic Chess Notation~\citep{hooper1996} is used to represent the chessboard and assess cognitive aspects.

\textbf{Perception.} To assess perception, we simulate a chessboard state by providing a sequence of moves. We then evaluate the system with questions of understanding the board state, including:
\begin{itemize}
    \item Understand chess piece position on a given FEN.
    \item Compute the number of captured pieces in Algebraic Chess Notation.
    \item Provide step-by-step analysis of pieces captured, the number of pieces left in total, and the number of pieces left for each player in Algebraic Chess Notation.
\end{itemize}
For the capture analysis questions, we reward partial understanding of the board. If a model manages to predict the number of captures partially, we will penalize only for the incorrect predictions, which include skipping piece capture or overestimating the number of captures. If the number of captures in the FEN is $n_c$ and the prediction from the model is $n_m$, the formula for scoring on this query is
\begin{equation}
\textit{Capture Analysis:} \quad s_{capture} = 1 - \frac{|n_c - n_m|}{n_c}\ .
\label{capture_analysis}
\end{equation}

Then, we get the perception score given FEN perception score $s_{FEN}$, capture analysis score $s_{capture}$, and piece analysis score $s_{piece}$ where each score is the number of questions with correct answers, 
\begin{equation}
\textit{Perception Score:} \quad s_{perception} = \frac{\text{$s_{FEN}$ + $s_{capture}$ + $s_{piece}$}}{\text{number of questions}}\ .
\label{perception_score}
\end{equation}

\textbf{Memory.} The system's memory is evaluated using questions that assess its general chess knowledge. Furthermore, the AI system is augmented with RAG~\citep{lewis2020retrieval} and has access to two chess books, simulating external knowledge sources. Questions specific to the system's external knowledge sources are included, requiring relevant sections and utilising the gathered context.

In this study, we investigate cognitive processes akin to long-term memory. As previously discussed, memory is encoded through a combination of the base model's knowledge and the RAG architecture. To enhance memory retention, we enable the system to incorporate new information into its vector database. Consequently, it can store relevant interactions with users, enriching its memory capacity.

The memory scoring questionnaire is designed to contain a singular solution, ensuring that the score is calculated by the formula as follows

\begin{equation}
    \textit{Memory:} \quad s_{memory} =
    \frac{\text{number of correct answers}}{\text{number of questions}}\ .
    \label{memory_score}
\end{equation}

This method facilitates an objective assessment of the memory performance.

\textbf{Attention.} 
The attention mechanism is subjected to a comprehensive assessment employing a tripartite approach. Initially, a sequence of chess moves is presented, accompanied by questions that require understanding specific segments of the chess moves. This evaluation aims to scrutinize the system's capacity for focused attention on pertinent data. Furthermore, the ability to answer any question and even participate in Q\&A tests demonstrates high attention quality when the system can identify the context of the question to use relevant knowledge. To test this, questions irrelevant to the control environment are introduced, assessing the system's ability to comprehend question context and access relevant information or acknowledge uncertainty. Notably, the accuracy of the response is secondary to the system's ability to recognise and utilise contextual cues. Finally, RAG questions are used to assess the system's ability to retrieve relevant context based on user queries. 
The questionnaire for attention contains a singular solution, ensuring that the score is calculated using Eq.~\eqref{memory_score}.

\textbf{Reasoning.}
To assess the system's reasoning capabilities, we have curated a collection of chess puzzles from \url{https://www.chess.com/puzzles}~\citep{chessdotcom}. We provide the initial chess board setup in FEN, which is then used to configure the board via the chess engine service. Ideally, our system should be able to select the appropriate service dynamically, which itself would demonstrate reasoning. However, for this proof-of-concept framework, we employ a query filtering script that detects keywords to trigger the chess engine service.
Once the board is set, the chess engine service attempts to solve the puzzle. The LLM's task is to generate explanations for the prediction of the best moves, which will be evaluated through human supervision using the rubric in Table~\ref{tb:rubric}.

% \begin{itemize}
%     \item[0] - Inadequate: The model exhibits solely erroneous assertions with no basis in reality. 
%     \item[1] - Deficient: The explanation contains elements of accuracy but is predominantly flawed. 
%     \item[2] - Satisfactory: The model correctly identifies the action taken but lacks sophisticated strategic insight. 
%     \item[3] - Competent: The explanation is free from false information and accurately describes the move.
%     \item[4] - Proficient: The model articulates a cogent rationale and strategic understanding behind the move.
%     \item[5] - Exemplary: The model provides an exceptional explanation that reveals a profound strategic acumen.
% \end{itemize}

\begin{table*}[!ht]
\centering
\setlength{\tabcolsep}{4pt}
\resizebox{\textwidth}{!}{
\begin{tabular}{cll}
\hline
{Score} & \multicolumn{1}{c}{{Notation}} & \multicolumn{1}{c}{{Explaination}} \\
\hline
0 & Inadequate & The model exhibits solely erroneous assertions with no basis in reality. \\
1 & Deficient & The explanation contains elements of accuracy but is predominantly flawed. \\
2 & Satisfactory & The model correctly identifies the action taken but lacks sophisticated strategic insight. \\
3 & Competent & The explanation is free from false information and accurately describes the move. \\
4 & Proficient & The model articulates a cogent rationale and strategic understanding behind the move. \\
5 & Exemplary & The model provides an exceptional explanation that reveals a profound strategic acumen. \\
\hline
\end{tabular}
}
\caption{\textit{Score scale for human study. This is an extension of the traditional five-point assessment framework commonly utilised in empirical research. For this study, a six-level scale has been employed to augment the granularity of differentiation among the evaluated models’ performances}}
\label{tb:rubric}
\end{table*}

For the system's reasoning quality, the LLM provides analysis for the $n^{\text{sys}}$ moves provided by the system solution, followed by a human evaluation using the six-level scale in Table~\ref{tb:rubric}.
The score for reasoning is then averaged over all the assessed analysis of the predicted moves on all $M$ puzzles,

\begin{equation}
    \textit{Reasoning:} \quad s_{reasoning} =
    \frac{1}{5 M}\sum_{i=1}^{M} \sum^{n^{\text{sys}}_{i}}_{k = 1} \frac{s_{i}(k)}{n^{\text{sys}}_{i}}\ .
    \label{reasoning}
\end{equation}

% \begin{center}
%    \begin{tabular}{|p{5cm}|p{3cm}|}
%     \hline
%     \textbf{Criteria} & \textbf{Contribution} \\
%     \hline Use the most appropriate service dynamically (\textit{in progress}) & Reasoning \\
%     \hline Explain the moves suggested by the chess engine service & Reasoning, Anticipation \\
%     \hline
%     \end{tabular} 
% \end{center}
% \textbf{Decision-Making/Problem-Solving:} To assess the system's decision-making/problem-solving capabilities, we use the same collection of chess puzzles curated from Chess.com~\citep{chessdotcom}. Each puzzle includes a predefined best-case solution, denoted as \( n_{\text{best}} \), representing the minimum number of moves required to solve the puzzle optimally. The system attempts to solve the puzzle by making a series of moves, with the total number of moves taken by the system denoted as \( n^{\text{sys}} \).

% For $j^{\text{th}}$ puzzle corresponding to $i^{\text{th}}$ quality which is \textit{decision-making}, the decision-making score of the agent to evaluate how fast the predicted solution with $n_{ij}$ moves can achieve the same goal compared with the $n^{\text{sys}}_{ij}$ moves provided by the system solution.
% It follows

% \begin{equation}
%     \textit{Decision:} \quad s_{ij} = \min(\frac{n_{ij}}{n^{\text{sys}}_{ij}}, 1)\ ,
%     \quad \forall j \in \mathcal{S}_i\ .
% \end{equation}
% \fi
%

% \begin{tabular}{|p{5cm}|p{3cm}|}
% \hline
% \textbf{Criteria} & \textbf{Contribution} \\
% \hline Solve chess puzzle using chess engine service & Decision Making \\
% \hline
% \end{tabular}

\textbf{Anticipation.}
A proficient chess player can predict their opponent’s moves, strategically capitalize on this foresight and plan several moves ahead. To assess this anticipation skill, we analyse the responses of an AI agent during puzzle-solving curated from Chess.com~\citep{chessdotcom}. 
% Specifically, we search for evidence of predictive reasoning in the agent’s decision-making process. 
% Additionally, the capacity to execute moves that compel the adversary to alter their planned actions—thus setting the stage for subsequent moves—serves as a clear demonstration of forward thinking and anticipation.

Each puzzle includes the best solution provided by Chess.com, denoted as \( n_{\text{best}} \), which represents the minimum number of moves to solve the puzzle. The system attempts to solve the puzzle by making a series of moves, with the total number of moves taken by the system is denoted as \( n^{\text{sys}} \).

% For $j^{\text{th}}$ puzzle corresponding to $i^{\text{th}}$ quality (which is Anticipation in the case), the anticipation score of the agent to evaluate how fast the predicted solution with $n_{ij}$ moves can achieve the same goal compared with the $n^{\text{sys}}_{ij}$ moves provided by the system solution.
For the anticipation quality, the LLM computes the ratio of the $n$ moves in the agent's prediction over the $n^{\text{sys}}$ best moves provided by the system, and the average score overall $M$ puzzles is

\begin{equation}
    \textit{Anticipation:} \quad s_{anticipation} = \frac{1}{M} \sum_{i=1}^{M} \min(\frac{n_{i}}{n^{\text{sys}}_{i}}, 1)\ .
\end{equation}

% =================================================================================================================================
\section{Experiments}

In this section, we present a full evaluation and analysis of the cognition performance of the proposed \opensi~AI System, which is featured by 5 cognition qualities: perception, memory, attention, reasoning, and anticipation.
We provide the quality scores on 5 LLMs in Fig.~\ref{fig:all_qualities}.
Our proposed AI system integrates LLMs with a query analyser and 4 services: a chess engine for best move prediction using Stockfish, a vector database for dynamic information retrieval, retrieval-augmented generation on documents, and code generation. The main LLMs used are GPT-4o, GPT-3.5 Turbo (for anticipation), Gemma 7B Instruct, Mistral 7B Instruct, and fine-tuned Mistral 7B.
All GPU-related experiments are conducted on NVIDIA GeForce RTX 3090, and our code is available at \href{https://github.com/TheOpenSI/CoSMIC.git}{https://github.com/TheOpenSI/CoSMIC.git}.

% ---------------------------------------------------------------------------------------------------------------------------------
\subsection{Evaluation on System Services}

\textbf{a) Query Analyser.}
The query analyser selects the best service using the LLM analysis of the semantic relevance between the query and all predefined services.
We generated 340 queries, including 40 queries for chess move prediction, 100 for vector database update with declarative sentences and documents, 100 for code generation, and 100 for general question and reasoning analysis, and achieved 82.63\% on Mistral 7B Instruct, 91.25\% on Gemma 7B Instruct, 95.75\% on GPT-3.5 Turbo, and 100\% on GPT-4o.
This provides an effective and automatic service routing on the query to our system.

\textbf{b) Best-move Prediction for Chess Game.}
The best-move prediction aims to predict the best next move for a given chess FEN or a sequence of moves.
Our system incorporates the strong open-source chess engine, Stockfish, with interaction with LLMs to analyse the move decision, yielding nearly perfect predictions to the ground truth labels obtained from \citep{chessdotcom}.
In Fig.~\ref{fig:all_qualities}, Gemma 7B Instruct, Mistral 7B Instruct, and fine-tuned Mistral are unable to predict any correct best moves, indicating their deficiency in the game reasoning. 

While GPT-4o alone exhibits an unexpectedly high prediction accuracy with a 32.5\% success rate, Fig.~\ref{fig:all_qualities}, it falls short in strategic reasoning, particularly in determining check and checkmate situations. In contrast, our system demonstrates superior performance by accurately predicting the optimal moves in 40 chess games, underscoring the significant benefit of integrating our chess engine service.

\textbf{c) Vector Database for Dynamic Information Retrieval.}
Similar to one-shot learning, our system can retrieve up-to-date context by adding certain information with an updated timestamp to the inbuilt vector database, which is managed by using the Facebook AI similarity search tool.
Information to be added can be sourced from a declarative sentence or a document in portable document format. 
%GPT-4o and Mistral 7B Instruct successfully update their answers with 100\% accuracy, demonstrating their ability to effectively incorporate new information into their responses.The fine-tuned Mistral 7B has 66.66\% success rate. In contrast, Gemma 7B Instruct achieves only 33.33\% success rate.

\textbf{d) RAG on Documents.}
In our system, the query into LLMs will first be used to retrieve relevant context from the vector database, followed by prompt generation under a tuned prompt template to trigger the LLM engine for text generation.
This retrieved-context will be filtered out if the score of its cosine similarity to the query is under a given threshold, 0.7 in our case, to avoid misleading information in the query to LLMs.
For information retrieval from an external document, our system uses LangChain to split the document into chunks with a chunk size 1,000 and overlap size 100, and then encodes them with an embedding model to update the vector database. The success rates achieved by GPT-4o, Gemma 7B Instruct, and Mistral 7B Instruct are  80\%, 70\%, and 77.5\%, respectively.

% ---------------------------------------------------------------------------------------------------------------------------------
\subsection{Evaluation on System Cognition Capability}

\textbf{a) Evaluation Scale.}
We evaluate the cognition ability of our \opensi~AI system with 5 qualities in Sec.~\ref{sec:introduction}.
The evaluation scale of each quality is provided below.

\begin{itemize}
    \item \textbf{Perception}. While the perception can be represented as the system's understanding of a scenario, we provide 3 types of questions: 40 questions for parsing chess piece position, 20 questions for parsing chess piece status, and 8 questions for identifying chess piece captures.
    \item \textbf{Memory.} The system's memory is evaluated on 3 types of questions: 40 questions for retrieving LLM's base knowledge, 8 questions for the memory of previous scenarios in a chess game, and 6 questions for retrieving context from the updated system vector database.
    \item \textbf{Attention.} We evaluate 40 RAG questions that are extracted or raised from 3 books. The correct attention should be localized on the page, providing the correct answer to the question.
    % \item \textbf{Reasoning.} The reasoning ability is measured by human study of 4 on the readability, precision, and logic of LLM's analysis of the best move prediction given the chess FEN. We provide 40 chess puzzles with each containing 1 to 3 sequence moves for the player of Black or White.
    \item \textbf{Reasoning.} The reasoning capability is evaluated through human annotation, assessing LLM's analysis of the suggested move by the chess engine service for a given FEN. A dataset of 40 chess puzzles is compiled, each comprising 1-3 move sequences for either Black or White. 4 human annotators rated the model's reasoning ability based on the rubric in Table \ref{tb:rubric}.
    \item \textbf{Anticipation.} With the same 40 chess puzzles used for reasoning, the anticipation ability is measured by the ratio of correctly predicted best moves using our system and comparison LLMs.
\end{itemize}

\begin{figure}[t]
\centering
% \begin{minipage}[b]{0.57\linewidth}
% \includegraphics[width=0.6\linewidth,valign=c]{figures/radar_plot.PNG}
% \hspace{0.02\linewidth}
% \includegraphics[width=0.35\linewidth,valign=c]{figures/test_2.png}
\includegraphics[width=0.9\linewidth]{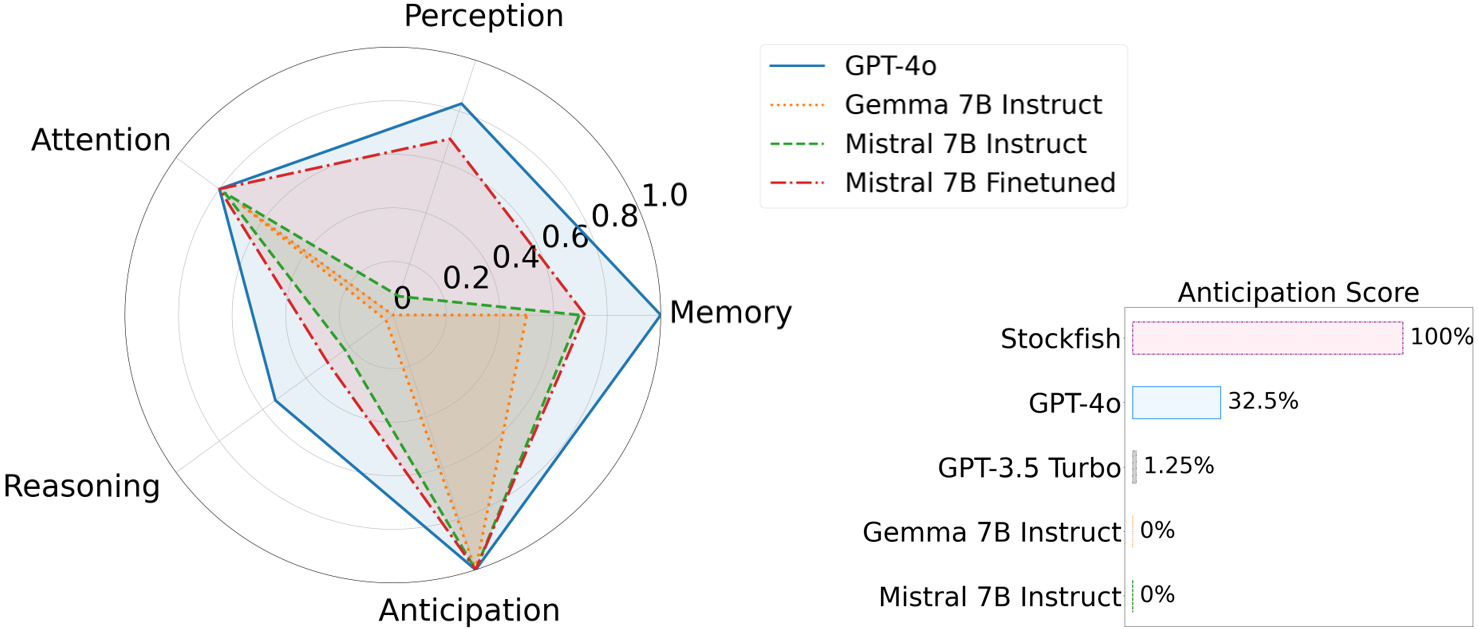}
\caption[Caption for figure]{The chart on the left displays the normalised scores for the five cognitive qualities measured using the \opensi~AI System powered with several LLMs. On the right, we evaluate the anticipation\footnotemark~for how well these LLMs can predict the next move in a chess game compared to our \opensi~AI System. LLMs are inferior in predicting the optimal next chess move.}
\label{fig:all_qualities}
% \end{minipage}
% \hfill
% \begin{minipage}[b]{0.4\linewidth}
% \includegraphics[width=\linewidth]{performance comparison.png}
% \vspace{10mm}
% \caption{\quad Anticipation Success Rate.}
% \label{fig:anticipation}
% \end{minipage}
\end{figure}

\textbf{b) System Cognition Capability.}
In Fig.~\ref{fig:all_qualities}, our system provides the statistics of 5 qualities evaluated on 4 public LLMs.
Particularly, GPT-4o performs the best, and Mistral 7B Instruct outperforms Gemma 7B Instruct on all qualities.
From the experiments, we observe that Gemma 7B Instruct is unable to support system prompts, leading to missing or irrelevant responses in several questions and its difficulty in adapting to varying evaluation scenarios and cognition qualities due to the rigid persona.
Different from the anticipation evaluated on base LLMs in Fig.~\ref{fig:all_qualities}, however, when embedding these models in our system with the anticipation operator changed to the inbuilt chess engine, all models achieve high anticipation scores by predicting the next best move given a puzzle FEN.

% We began by evaluating the two base models, Gemma 7B and Mistral 7B. Our findings indicated that Mistral 7B consistently outperformed Gemma 7B across all cognitive qualities. One significant limitation observed with Gemma 7B was its lack of support for system prompts, which made it refuse to provide an answer in several test cases. Additionally, Gemma's rigid persona prevented it from adapting to different tests and qualities, resulting in poor overall performance.

%The fine-tuned Mistral 7B exhibited marked improvements across the board

% Among the models, GPT-4 scored the highest in reasoning by explaining the rationale behind its suggested next move. To delve deeper into predictive capabilities, we conducted an additional experiment where GPT-4 was tasked with predicting the next move without the aid of the chess engine service. GPT-4 managed to predict the next optimal move correctly only 23.33\% of the time. While GPT-4 could often suggest the correct next move, it rarely managed to accurately anticipate the resultant board state after the move. \Adnan{did not mention the result when we don't consider + or \# sign}

Among these LLMs, GPT-4o emerges as the frontrunner in reasoning, articulating the logic underpinning its subsequent actions.
% A particular investigation into its proficiency in predicting subsequent chess moves without the chess-engine service, giving a 32.5\% success rate on the 40 reasoning questions, is its success in precisely understanding the chess board FEN.
These underscore the potential of our \opensi~AI system to augment the capabilities of such LLMs through the embedded services.
For instance, the integration of GPT-4o and the chess engine has shown an improvement in both reasoning and anticipation through the \opensi~AI system. Furthermore, given the superior performance of  Mistral 7B Instruct, we fine-tuned Mistral 7B model to enhance its cognitive capability for domain-specific tasks.

\footnotetext{We evaluated GPT-3.5 Turbo and GPT-4o from OpenAI on chess tasks and found that GPT-4o demonstrated superior chess understanding. Hence, we use GPT-4o to evaluate the system's cognition ability as a strong comparison.}
\section{Future Work}
In our current system, the query analyser utilises keyword detection to route user queries to appropriate services and employs a Chain of Thought (CoT) query service to convert user queries into CoT queries through keyword matching. While this method has proven effective in enhancing the system's cognitive abilities, it presents certain limitations in terms of flexibility and scalability. To address these challenges, we propose improving the query analyser by transitioning from the keyword-based detection mechanism to a fine-tuned LLM or a classification model. This enhancement will enable more accurate, context-aware routing of user queries to relevant services. Additionally, we plan to automate and refine the CoT query generation process by training an LLM to generate CoT queries, further improving the system’s reasoning capabilities~\citep{jin2024impact}.

As part of our ongoing efforts to advance the system, we are currently working on a self-correcting mechanism for the software development workflow. This feature will allow the AI to identify and correct errors during the code generation process, aiming to increase both efficiency and reliability. We are benchmarking this capability against similar approaches to ensure its effectiveness in improving overall system robustness and adaptability in real-time scenarios.

To further enhance the system’s scalability and reliability, we aim to replace the single LLM at the core with a collection of fine-tuned LLMs, each specialized for different domain-specific tasks. These models will be deployed within a distributed system architecture, allowing the system to handle a higher volume of queries without compromising cognitive performance. This approach will reduce computational complexity, streamline system maintenance, and facilitate seamless updates and the integration of new services. Furthermore, the distributed architecture will make the system adaptable to consumer-level computing resources, such as GPUs, allowing the deployment of specialized components to address a wide range of tasks across different domains.

% =================================================================================================================================
\section{Conclusion}

In this study, we have showcased the efficacy of integrating multiple AI systems with LLMs to augment the cognition abilities of digital assistants. Our proposed architecture is resilient and intuitive, allowing for seamless incorporation into broader systems. One of the core innovations lies in the query analyser for specific services, thereby enhancing the LLM’s role as an interactive intermediary between the user and the system. Furthermore, we have illustrated that while LLMs may exhibit limited predictive capabilities, they are competent at interpreting the responses from predictive tools. This enhanced system holds the potential to assist a wide spectrum of professionals, including financial advisors, lawyers, programmers, etc.
In summary, the proposed system integrates multiple AI models and services to create a cohesive framework to demonstrate comprehensive cognitive abilities. By integrating these components, our system aims to bridge the gap between raw computational abilities and human-like cognitive processes, setting the stage for future advancements in AGI.
% =================================================================================================================================
\ifarxiv
    \bibliographystyle{abbrvnat}
\else
    \bibliographystyle{misq}
\fi

\bibliography{bibliography}

% =================================================================================================================================
\section*{Acknowledgments}
This work is funded under the agreement with the ACT Government, Future Jobs Fund - Open Source Institute (OpenSI) - R01553; and  NetApp Technology Alliance Agreement with OpenSI - R01657. Buddhi Gamage's research is supported by a NextGen Robotics - University of Canberra co-funded scholarship. 
%We especially acknowledge Carlos C. N. Kuhn for their visionary approach and foundational ideas that made this research possible.

%=================================================================================================================================
\section*{Author Contributions}
Muntasir Adnan and Zhiwei Xu collaborated on implementing the code base, designing experiments, evaluating results, and writing the manuscript.
Buddhi Gamage focused on experiment design, result evaluation, and manuscript writing.
Damith Herath conducted a thorough review of the manuscript.
Carlos C. N. Kuhn developed the initial concept for this project and contributed to experiment design, research question development, data evaluation, and manuscript writing.

%=================================================================================================================================
\section*{Copyright}
\textbf{Copyright} © 2024 OpenSI. This is an open-access article licensed under \href{https://creativecommons.org/licenses/by-nc/4.0/deed.en}{a Creative Commons Attribution-Non-Commercial 4.0 Australia License}, which permits non-commercial use, distribution, and reproduction in any medium, provided the original author and ACIS are credited.

\end{document}